# Improving automatic detection of driver fatigue and distraction using machine learning

By

Dongjiang Wu

Student ID:  2499741

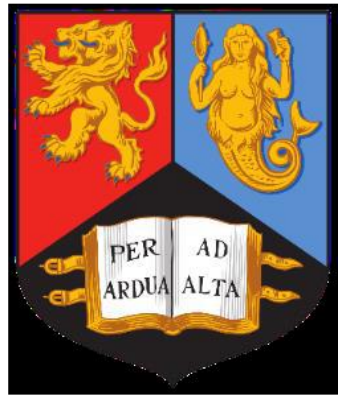

Supervisor: Dr Ruchit Agrawal

A thesis submitted to the University of Birmingham
For the MSc degree in Computer Science and Artificial Intelligence

School of Computer Science
University of Birmingham
Birmingham, Dubai

September 2023

# Contents

**Abstract**

Changes and advances in information technology have played an important role in the development of intelligent vehicle systems in recent years. Driver fatigue and distracted driving are important factors in traffic accidents. Thus, on-board monitoring of driving behaviour has become a crucial component of advanced driver assistance systems (ADAS) for intelligent vehicles. In this article, we present techniques for simultaneously detecting fatigue and distracted driving behaviors using vision-based and machine learning-based approaches. In driving fatigue detection, we use facial alignment network (FAN) to identify the 68 facial feature points in the image [32]. Calculating facial feature points' distance to detect the opening/closing of the eyes and mouth. In distraction detection, we use a convolutional neural network (CNN) based on the MobileNet architecture to identify various distracted driving behaviors. Experiments are performed on the PC with a webcam and using public datasets and our own datasets for training and testing. Compared to previous approaches, we build our own datasets and provide better results in terms of accuracy and computation time.

*Keywords*: *FAN, Open CV, Dlib, CNN, MobileNet, PERCLOS, fatigue detection, drowsiness detection, distraction detection*

# *Acknowledgment*

I would like to express my sincere appreciation to my supervisor Dr. Ruchit Agrawal for his guidance and support during the project. This research would not have been completed without his mentorship and advice during the semester. Also, I would like to thank my wife for her continued encouragement, support and patience. They are the main source of inspiration and creativity in my life.

# 1    Introduction

## 1.1    Motivation

According to statistics from the World Health Organization (WHO), road accidents claim the lives of millions of people every year. The American Automobile Association reported that 7% of all crashes and 21% of fatal traffic crashes are due to driver fatigue [1]. And the U.S. National Highway Traffic Safety Administration (NHTSA) reported that 8% of fatal crashes, 14% of crashes with injuries, and 13% of all police-reported motor vehicle crashes in 2021 were caused by distracted driving [2]. Given these statistics, driver fatigue and distracted driving have gradually become the leading causes of road accidents. Fatigue impairs the driver's senses and decision-making abilities when it comes to controlling a vehicle [9]. Some traffic laws also prohibit driving for a long time. Therefore, detecting both driver fatigue and distracted driving is essential to improve road safety.

Many methods and experiments have been applied for the detection of fatigue driving. According to different input characteristics, driver fatigue detection can be divided into three categories: physiological parameters, vehicle data and facial features.

1)    The physiological approach, such as brain waves (using electroencephalogram - EEG) [3], [4], heart rate (using electrocardiogram - ECG) [5] and electrical signals from muscle cells (using electromyogram - EMG) [6], are generally reliable for detecting drowsiness because its correlation with the driver's body alertness is relatively strong. However, it has a limitation due to the physical connection on the driver's body, it will interfere somewhat with driving behavior [10].

2)    Approach to vehicle data, using both proprioceptive and exteroceptive vehicle driving data, such as steering wheel, driving speed, braking, acceleration and level crossings, recorded by the on-board diagnostic system [7]. This approach can be error-prone due to road conditions, driving style and vehicle characteristics.

3) Facial features approach, using computer vision techniques to monitor driver condition by analyzing facial expression characteristics, eye closure, and mouth condition in real time.

For distracted driving, when the driver engages in other activities while driving, such as texting or talking on the phone, talking to other passengers, eating or drinking, etc., will reduce the attention paid to driving. With the computer vision approach, this becomes possible by analyzing the driver's head pose [8].Three main approaches for driver fatigue and distraction detection are compared in Table 1-1 [33].

| | Approaches based on physiological approach | Approaches based on vehicle data | Approaches based on facial features |
|---|---|---|---|
| Fatigue Detection | Yes | Yes | Yes |
| Distraction Detection | No | Yes | Yes |
| Accuracy | Very Good | Good | Moderate |
| Simplicity | Difficult | Relatively Easy | Easy |
| Detection Speed | Very Fast | Slow | Fast |

Table 1-1, Comparison between the three main approaches for driver fatigue and distraction detection.

## 1.2    Structure of the report

This report is divided into three main sections: introduction, literature review and background, as well as data collection and processing, design network, results and evaluation, discussion, conclusions and perspectives.

The first part is the introductory section, which includes two sections on the motivation and structure of the report. Giving the motivation, motivation and purpose of this study will be exposed, explaining why fatigue and detection of distracted driving are of great importance in modern society. Then, in the structure of the report, the organizational framework of the entire report will be described, and the content and objectives of each chapter will be clarified in order to provide a general guide to readers.

The second part is a literature review, which includes several sections such as computer vision in fatigue detection, computer vision in distraction detection, application of vision and machine learning in driving behavior detection, data sets and evaluation indicators, and research progress and challenges. By reviewing the literature in related fields, the current state of research and methods for detecting fatigue and distracted driving, as well as existing issues and challenges will be presented.

The third part is the background, which includes sections on computer vision, Perclos, CNN and MobileNet. In this section, the basic knowledge and techniques closely related to this research will be presented, including the fundamentals of computer vision, the application of Perclos (percentage of eye closure) as an indicator of fatigue, the characteristics and applicability of network architectures such as CNN (convolutional neural networks) and MobileNet.

Through the organization of the three parts above, this report will comprehensively present the context of the research, the review of relevant literature and key contents such as data collection and processing, network design, results and evaluation, in order to provide a clear structural framework for readers to better understand and read the detailed content of the following chapters.

## 2    Literature review

### 2.1    Computer vision in fatigue detection

In recent decades, a number of different methods for detecting driver drowsiness have been proposed. Yaman et al. [31] has classified four methods according to different research directions. Figure 1 shows all the latest used methods for classifying driver drowsiness levels. The first two are from driver perspective: image and biological based. The third one is from the vehicle itself. The fourth one is hybrid methods, combined comes from previously mentioned ones.

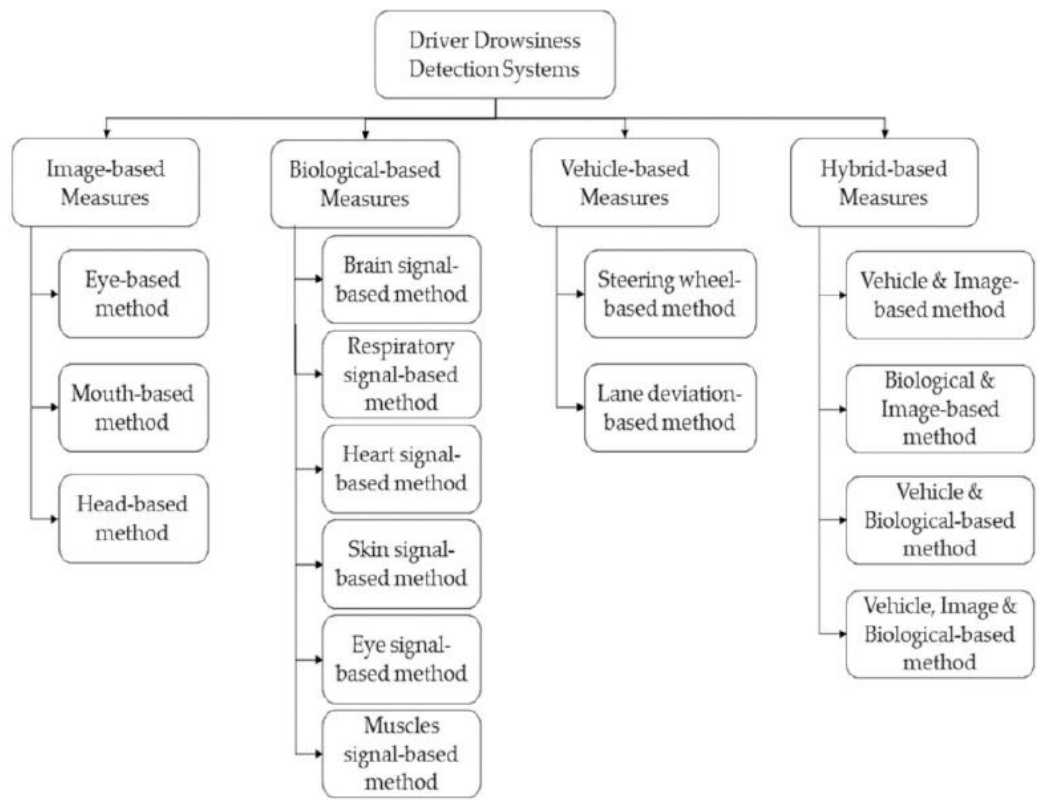

Figure 2-1, Driver drowsiness detection measures

Most Fatigue detection methods rely on driver's behavior characteristics refers to using the variation of driver's face expression, eyes, mouth status to determine the driver's fatigue level. Eye status are considered as one of the most important feature on reflecting driver's drowsiness. "PERCLOS" (the percentage of eye closure over time) for fatigue measurement was first propose by Wierwillle [10] in year 1994, using 'eye' closing time to reflect the fatigue state.

In addition, vision and machine learning are widely used in detecting driving behavior to problems such as drowsy driving and distracted driving. By analyzing information such as the driver's facial expressions, eye movement data and physiological signals, the driver's fatigue state and attention distribution can be accurately identified. This approach often combines computer vision technology and machine learning algorithms to be highly feasible and in real time.

## 2.2    Datasets and evaluation measures

In behavioural detection research, the selection of datasets and the design of assessment indicators are very important. A good data set can provide various driving scenarios and examples of real-world driving behavior, which effectively evaluates the algorithm's performance and generalization ability.

At present, some publicly driving behaviour datasets are already available to researchers. For example,  YawDD dataset, which contains 322 male and female drivers for the fatigue driving detection . Some examples images are shown in Figure 2-2:

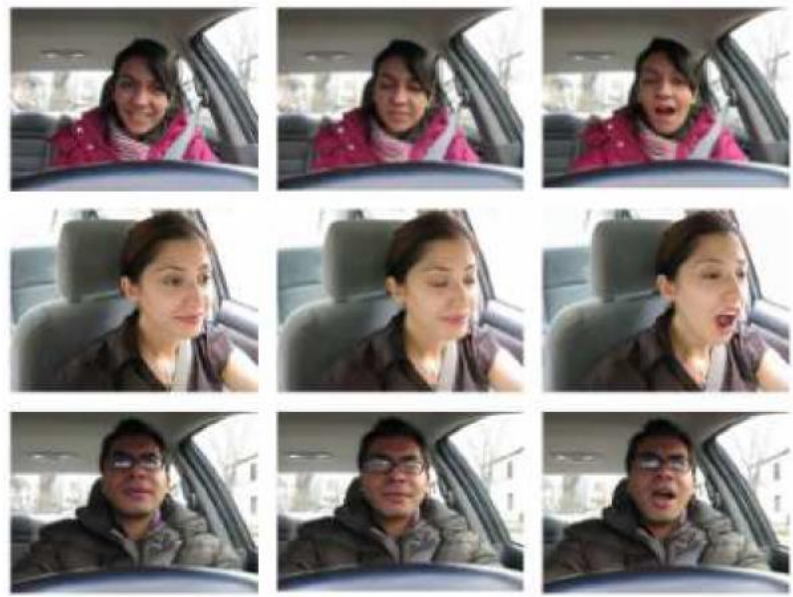

Figure 2-2, some example images in YawDD dataset

The SFDDD(State Farm Distracted Driver Detection) dataset, captured by State Farm for Kaggle driving distraction competition, and they set up these experiments in a controlled environment - a truck dragging the car around on the streets - so these "drivers" weren't really driving. There are totally captured 22,424 images classify into 10 classes, such as 'safe driving'(c0), 'texting-right hand' (c1), 'talking on the phone-right hand' (c2), 'texting-left hand' (c3), 'talking on the phone-left hand' (c4), 'operating the radio' (c5), 'drink' (c6), 'reaching behind' (c7), 'hair and makeup' (c8), 'talk to passengers' (c9). These images were captured from right side of driver by a camera which installed at passenger visor. Some examples images are shown in Figure 2-3:

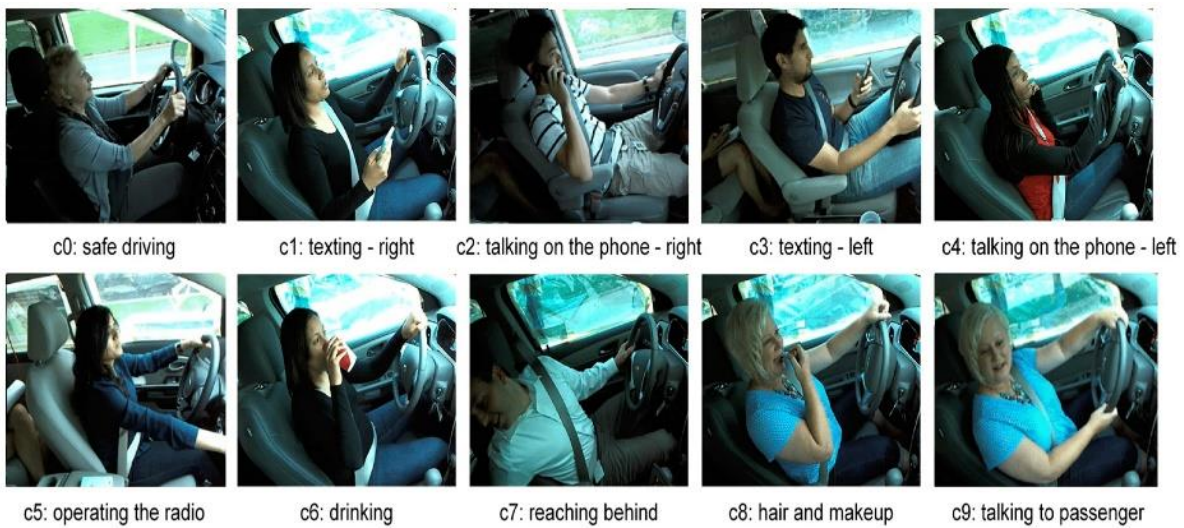

Figure 2-3, example images from 10 classes of State Farm Distracted Driver Detection

These datasets provide a multitude of training and test samples that can be used to develop and evaluate different algorithms for detecting driving behavior. In this project, we want to use one camera to do two tasks in simultaneously for fatigue and distraction detection apart from SFDDD, we designed five classes dataset and collected with a front camera that invests time and effort. Some examples images are shown in Figure 2-4:

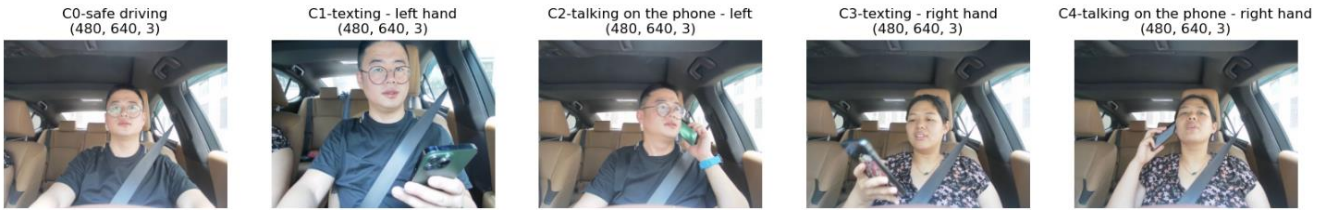

Figure 2-4, example images from 5 classes of private FrontView dataset

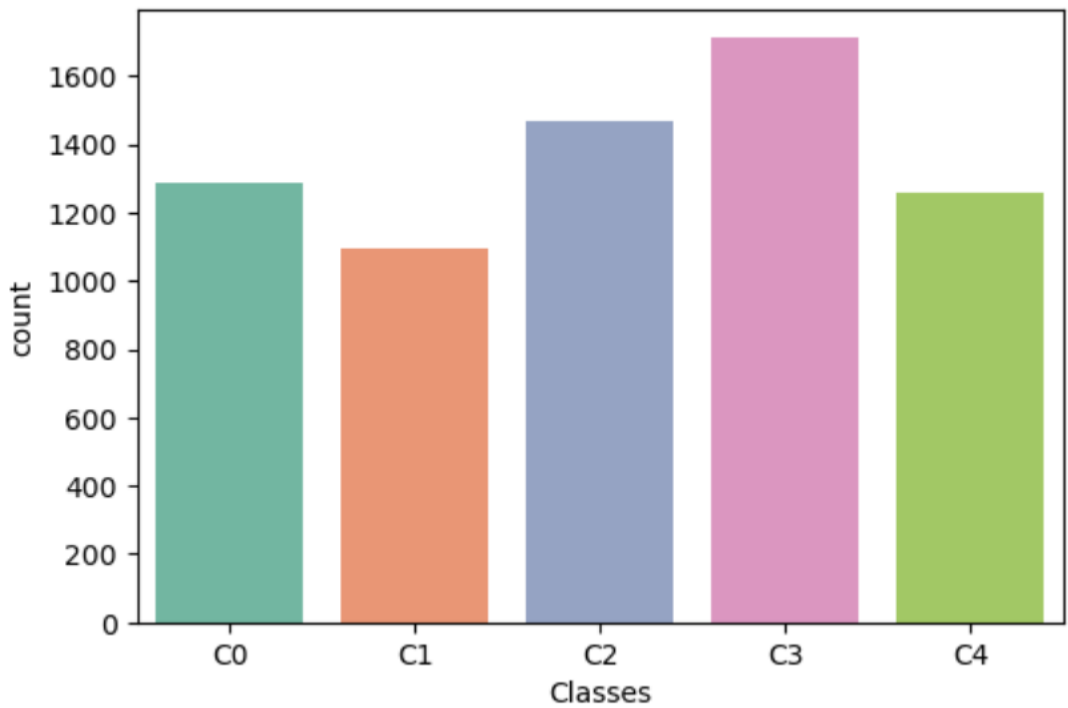

Figure 2-5, images distribution of FrontView dataset over 5 classes

In terms of assessment measures, measures include accuracy, precision, recall and F1 score. Precision refers to the ratio between the number of samples correctly classified by the classifier and the total sample size.

Future research could consider designing more nuanced and rigorous evaluation methods to more accurately assess and compare the performance of different algorithms for detecting driving behaviour.

**2.3    Research progress and challenges**

Reading the findings of the literature of many researchers, significant progress has been made in the study of driving behavior detection, but there are still challenges to be solved.

First, with the development of deep learning technology, vision-based driving behavior detection and machine learning has made significant breakthroughs. Deep learning models such as convolutional neural networks (CNNs) and recurrent neural networks (RNNs) can extract rich functionality from image and video data and achieve efficient and accurate classification of driving behavior. These methods not only improve classification performance, but also adapt to changes in different scenarios and driving environments.

Second, researchers are also beginning to explore the use of multimodal data. In addition to image and video data, other sensor data (e.g. radar, inertial measurement units) and vehicle information (e.g. speed, acceleration) can be combined to obtain more comprehensive information about driving behaviour. By merging multiple data sources, the robustness and accuracy of driving behavior detection can be improved. Significant improvement has been demonstrated across various language and vision tasks using the incorporation of contextual and multi-modal information [11-20].

However, a number of challenges remain. First of all, driving behavior is a complex issue, influenced by many factors, such as individual differences between drivers, road conditions and traffic environment. It is difficult to construct more nuanced and accurate models to reflect the relationship between these factors. Actual data on driving behavior is often scarce and unbalanced, making the process of collecting and labeling data difficult. How to solve the problem of insufficient data and adopt effective methods of augmenting and balancing data to improve the generalization capability of algorithms is the direction that requires further research.

In addition, applying driving behavior detection to real-world driving scenarios presents some challenges. For example, in the field of autonomous driving, how to combine the detection of driving behavior with decision-making to achieve the safety and reliability of intelligent driving systems is still a problem to be solved.

In summary, while there have been significant advances in detecting driving behaviour, there are still challenges to overcome. Future research can be devoted to exploring the complexity and diversity of driving behaviour and designing more robust and effective approaches to address the challenges of real-world driving scenarios.

## 3   Background

### 3.1   Computer vision

Computer vision is the study of how computers acquire the ability to understand and analyze images and videos [21]. With the continuous advancement of computer technology, computer vision technology has made major breakthroughs in many fields and has produced a wide range of applications.

First, computer vision relies on digital image processing technology, which processes and analyzes images in a computer by converting analog image data captured by the sensor into a digital image. These processes include image enhancement, noise reduction, edge detection, etc., to extract features and information from the image and provide accurate input data for subsequent visual tasks.

Second, computer vision uses machine learning and deep learning techniques to solve complex vision problems by training algorithms with self-learning capabilities. These algorithms can learn the patterns, structure, and semantic information of the image by forming a large amount of labeled image data, and then can identify, classify, locate unknown images, and other operations. In this project, the method of using facial features and CNNs of convolutional neural networks in the detection of driver fatigue and distraction is a technology of this machine learning and deep learning [22].

Finally, computer vision also relies on theories and methods in related fields such as pattern recognition and computational geometry to perform tasks such as detecting, tracking, and segmenting objects in images. Pattern recognition technology can extract images and identify target objects with specific features or shapes, such as driver fatigue detection, recognizing the open/closed state of the eyes and mouth. Computational geometry, on the other hand, involves modeling and analyzing geometric structures and spatial relationships in images to obtain accurate positioning and estimation of target exposure [23].

In summary, computer vision, as an interdisciplinary discipline, is committed to equipping computers with the ability to understand and analyze images based on digital image processing, machine and deep learning, and pattern recognition. Its application in the field of fatigue detection and driver distraction has a wide range of perspectives and can provide effective assistance and guarantee for safe driving.

## 3.2    Perclos

Perclos (PerClos) is a commonly used method in the detection of fatigue at the wheel, which assesses the level of drowsiness and fatigue by analyzing the condition of the driver's eyes, thus reminding the driver to take the necessary measures to avoid accidents.

The Perclos method is based on the characteristics of a reduced blink frequency and a longer eye closure time in the human eye under fatigue. By using a device such as a camera or infrared sensor to monitor the condition of the driver's eyes, Perclos can calculate the percentage of eyelid closure over a period of time. When the closing time of the eyes exceeds a certain threshold, the driver may be judged to be in a state of fatigue or drowsiness and reminded to take the necessary rest measures [24].

More precisely, Perclos is calculated as follows: Perclos = (Eye closing time / Total time) × 100%. In practice, the closing time of the eyes is defined as the sum of the time intervals between two consecutive closures of the eyelids and a threshold is set, such as 20% or 30%. If the calculated Perclos value exceeds the defined threshold, it means that the driver has a high level of drowsiness and must pay attention to safety.

The Perclos method has many advantages. First of all, it is based on direct observation of the condition of the eye and is relatively easy to implement and use. Second, Perclos can capture driver fatigue and drowsiness in time, providing early warning and opportunities for intervention before accidents occur. In addition, Perclos can be combined with other sensors or monitoring methods, such as vehicle speed, steering wheel position, etc., to improve the accuracy and reliability of detection.

However, the Perclos method also has some limitations. First, it can't distinguish between the driver's natural blinking and the closing behavior of the eyes caused by fatigue, which can lead to false positives. Secondly, the Perclos method is more demanding on the camera and requires reliable surveillance at night or in low light conditions. In addition, individual driver differences and environmental factors can also impact Perclos' accuracy [25].

In summary, Perclos, as a method of detecting fatigue based on the condition of the eye, evaluates the level of fatigue of a driver by calculating the percentage of eyelid closure. Despite some limitations, Perclos has achieved some results in practical applications and can be combined with other detection methods to improve the safety of drowsy driving.

## 3.3   CNN

The convolutional neural network (CNN) is a deep learning model applied to image recognition and computer vision tasks. By simulating how the human visual system works, CNNs can automatically extract features from input images, classify and recognize them. In this topic, driver fatigue and distraction detection, CNNs are widely used to analyze and process data on driving behavior [26].

Convolutional neural networks consist of several layers, including convolutional layers, clustering layers, and fully connected layers. Among them, the convolutional layer extracts local features from the image through a series of convolution operations to capture detailed information about the driver's face or behavior. The pooling layer is responsible for reducing the size of the feature map and preserving the most important features. Finally, the fully connected layer classifies and discriminates extracted entities [27].

Using CNNs based on the MobileNet architecture, various distracted driving behaviors can be effectively identified [27]. The architecture aims for a lightweight design while maintaining good performance. By feeding images or video data of driving, CNNs can learn and extract key characteristics, including distracting behaviors such as cell phone use, feeding, and yawning. By training large-scale datasets, CNNs can accurately determine whether drivers are behaving distracted in different scenarios, providing support for warnings and subsequent interventions.

Creating your own dataset can provide better results than traditional methods. By collecting large amounts of data on actual driving behavior, including fatigue and distraction behavior, CNN models can be trained and tested more accurately. Self-constructed datasets can not only cover more behavioral situations and driving scenarios, but also address the biases and limitations of some public datasets. In addition, self-built datasets can also help optimize CNN models, improve accuracy, and reduce computation time for better application in real-world driving scenarios [28].

Convolutional neural networks (CNNs) are one of the techniques commonly used in vision-based and machine learning-based methods to detect driver fatigue and distraction. By designing appropriate architectures and leveraging self-constructed datasets, CNNs can effectively identify fatigue and distracted driving behaviors and provide drivers with timely warnings and interventions. With the continuous optimization of algorithms and the improvement of data sets, CNNs will have a broader application perspective in the field of driving behavior analysis.

## 3.4    MobileNet

MobileNet is a lightweight convolutional neural network architecture specifically designed to enable efficient image classification and object detection on devices with limited computing resources. It adopts the idea of deep separable convolution to separate traditional convolution operations into two stages: deep convolution and point convolution, which significantly reduces the number of parameters and the amount of computation [29].

By introducing the concept of deep separable convolution, MobileNet offers greater computational efficiency and smaller model size, making it widely used in resource-constrained scenarios such as embedded devices and mobile devices [30]. Compared to traditional convolutional neural networks, MobileNet significantly reduces the number of parameters and calculations for the model while maintaining high accuracy, allowing it to operate in real time with lower hardware configurations.

In this study, the convolutional neural network based on the MobileNet architecture is selected as a network model for detecting fatigue and distracted driving [31]. Since recognition of driver fatigue and distracted behavior must be performed in a real-time environment, an efficient and accurate model is needed to process real-time image streams. MobileNet's lightweight nature makes it ideal for use on mobile devices and embedded platforms, and it has shown good performance in our experiments [32].

Using the convolutional neural network based on the MobileNet architecture, this project can effectively detect driver fatigue and distracted behavior [33]. Compared to previous methods, this project provides more accurate and efficient results by building its own dataset and training and optimizing the network. The choice of MobileNet not only meets our computational resource constraints [34], but also guarantees the performance and reliability of the model, making an important contribution to the development of driver fatigue and distraction detection technology.

## 4 Data collection and processing

### 4.1 Data collection methods

#### *4.1.1 Data acquisition equipment*

Data acquisition devices are an essential part of detecting driver fatigue and distraction, acquiring data on the features and behavior of the driver's face to provide the basis for further data processing and analysis [35].

The data collection method in this project is to use a webcam that can be easily connected to a computer or mobile device and capture the driver's facial features and behavior. By placing a camera inside the car, such as the dashboard or windshield, a live video feed from the driver can be obtained. Such a video stream can be used to analyze facial features, such as opening and closing eyes and mouth.

#### *4.1.2 Design and implementation of data acquisition processes*

The design and implementation of a webcam data acquisition process is a critical step in a fatigue and distracted driving detection study, starting with the selection of a suitable webcam for the study. Consider factors such as camera resolution, frame rate, and light sensitivity to ensure a clear and stable video stream [36]. Then, depending on the research needs and the actual situation, select the appropriate installation position inside the car. Common mounting locations include above the dashboard or windshield to be able to capture the driver's facial features. After that, set the video capture parameters of the camera, including resolution, frame rate, etc. Depending on the limitations of hardware devices and computing resources, appropriate adjustments are made to ensure the quality of the captured video stream and meet real-time requirements.

In the real-world driving scenario, the driver's video data is recorded. Make sure the camera captures the driver's facial features in its entirety and set the appropriate recording time to get enough data on driving behavior. At the same time, the collected video data is annotated and annotated to provide training data for supervised learning. For example, the condition of facial features (opening/closing of eyes and mouth) in each frame is labeled and labeled for classification or regression. Finally, establish a reliable data storage and management system to ensure that collected data is properly preserved and organized. This can include creating folder structures, naming conventions, and backup and recovery policies.

However, during the data collection process, it is guaranteed that the driver's privacy is fully protected. Comply with laws and regulations, ensure that data is only used for research purposes, take necessary security measures, such as encrypted transmission and storage, and regularly check the quality of the data collected, including the clarity, stability and integrity of video streams. Resolve possible issues such as blurry images, lost images, or other distractions.

### 4.1.3   Environmental media and conditions for data collection

When collecting data, the choice of environmental parameters and conditions is important for data quality and accuracy. The first step is to ensure that the data collection environment has the right lighting conditions to achieve clear images and videos. Environments that are too dark or too bright can result in clear capture of facial features or overexposure of images, so proper control of indoor or outdoor light is essential [37].

Second, reduce factors that can interfere with data acquisition. For example, avoid interference between strong background light sources, reflective objects or other obstacles, and the driver's face.

During the data collection process, it is possible to consider simulating the actual driving environment, including road conditions, traffic conditions, etc. This allows for a more realistic reflection of the driver's behaviour and reactions, while ensuring that data recording devices, such as cameras, are securely attached to the selected mounting position, thus avoiding jolts or movements while driving that could affect the quality of data collection.

Keep drivers engaged and focused during the data collection process. Drivers should drive by normal standards and try not to be affected by additional distractions or negative emotions.

Finally, in order to obtain reliable data, it is recommended that any technical problems or anomalies be monitored, reported and addressed in a timely manner. This improves the quality and reliability of data collection and effectively supports further data analysis and algorithm development.

## 4.2    Data pre-processing

### 4.2.1    *Data cleansing and denoising*

In data pre-processing for driver fatigue and distraction detection, data cleaning and denoising helps improve data quality and reduce noise and aberrant interference on model training. Data cleansing aims to solve problems such as missing values, outliers, and duplicate values in datasets, making datasets more reliable and consistent.

First, verify that the dataset contains missing values and process them accordingly. You can choose to delete samples/features that contain missing values or perform fill operations such as medium fill, interpolation padding, and so on. we did data cleaning from SFDDD training dataset, for c0 to c4 classes. For example, below two images from c1 class, which has abnormal occlusion on it.

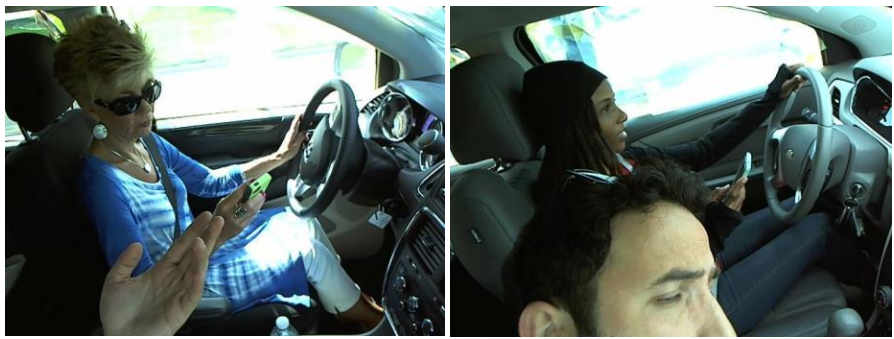

Figure 4-1, clean-out images example from SFDDD dataset

For outliers, identify and process outliers using statistical methods such as outlier detection algorithms, visualization methods, or domain knowledge. Outliers can be excluded based on thresholds or corrected using methods such as interpolation. For duplicate value processing: Detect and remove duplicate samples from the dataset to avoid duplicate effects on model training.

Data denoising aims to reduce noise in the dataset, improve the model's ability to model real signals, apply high-pass filters to filter out low-frequency noise, and preserve details and edge information in the image. This project is implemented by frequency domain methods (such as Fourier transforms) or spatial domain methods (such as Laplace operators), applying low-pass filters to smooth the image and reduce high-frequency noise. This eliminates subtle fluctuations and image noise. Commonly used methods include medium filtering, Gaussian filtering, etc. The data is then smoothed by applying smoothing techniques such as moving average, median filtering, to eliminate peaks and noise from the data.

Through data cleansing and denoising processing, this project eliminates noise and outliers from the dataset, making it more reliable and consistent. This will provide more accurate and effective results for further training and testing of the model.

### 4.2.2 Data calibration and alignment

Data calibration and alignment are important steps in data preprocessing in detecting driver fatigue and distraction. These steps help ensure the accuracy and consistency of facial features and behavioral information in your dataset.

Data calibration aims to adjust and normalize facial features in a data set to eliminate differences introduced due to factors such as pose, lighting, and camera angle changes under different conditions. By detecting the position of key points on the face, such as the eyes and mouth, posture calibration can be performed based on their relative position and angle.

The second is scale calibration, using a known reference scale, such as a fixed width or height of a face, to scale facial features in the dataset so that they have a consistent size.

Lighting calibration is then performed to adjust the brightness and contrast of the image to reduce image differences due to changes in lighting conditions. Image enhancement techniques such as histogram equalization, gamma correction, and other image enhancement techniques can be used to manage lighting variations.

Data alignment aims to align the facial features of a dataset to a unified coordinate system to ensure position relationships and spatial consistency across features. Use rigid body transformations (e.g., rotation, translation) to align facial features with a reference coordinate system. This can be achieved by calculating the relative distance and angle between facial features. Use image recording techniques to align images in a dataset to a reference or medium image. This can be achieved through feature extraction and matching, which uses feature descriptors and recording algorithms such as SIFT and SURF. Depending on the input requirements of the model and algorithm used, images in the dataset are cropped to remove irrelevant areas or retain regions of interest, and in this topic, only areas containing facial features are preserved.

With data calibration and alignment processing, you can eliminate differences due to changes in posture, scale, and lighting, and ensure accurate and consistent facial features and behavioral information across your dataset. This will provide more reliable and accurate results for subsequent training and testing, and improve the performance of driver fatigue and distraction detection systems.

### 4.2.3   Data sampling and labelling

Data sampling is one of the important steps in data preprocessing, which involves selecting the appropriate data sample to represent the data set. In detecting driver fatigue and distraction, proper sampling of data can help us obtain representative and diverse data, which can improve the generalizability and accuracy of the model.

First, in the data sampling process, this project ensures that the dataset contains a variety of possible driving conditions and scenarios, such as day, night and driving scenarios under different lighting conditions. This helps train the model to effectively detect fatigue and distraction in various real-world situations.

Second, data from different drivers is sampled to cover different driving styles and individual differences. There may be differences in the facial features and behavioral habits of different drivers, and the generalization of the model can be improved by sampling data from multiple drivers.

In addition, in order to better capture facial features, high-resolution images were used for sampling. This can be achieved by adjusting the camera settings or using a higher resolution camera.

In fatigue tests, this section marks open/closed eye and mouth conditions and uses binary labels (closed/non-closed) to indicate eye and mouth condition.

In distraction detection, this section identifies various distracted driving behaviors, namely using a mobile phone, setting the radio, looking in the car, making a phone call and assigns a multi-category label to each sample to indicate the distracted behavior of the current driver.

When labeling data, this topic uses automated algorithms to make preliminary annotations, and then I check and correct them to ensure the accuracy and consistency of the labels.

Through reasonable data sampling and accurate data annotation, this project successfully builds a rich and reliable data set, which provides better training and testing data for vision- and machine learning-based fatigue and driver distraction detection technology. Thus improving the accuracy and performance of the model.

The model of the flat vector collection process that converts an image to an accentuated edge is as follows:

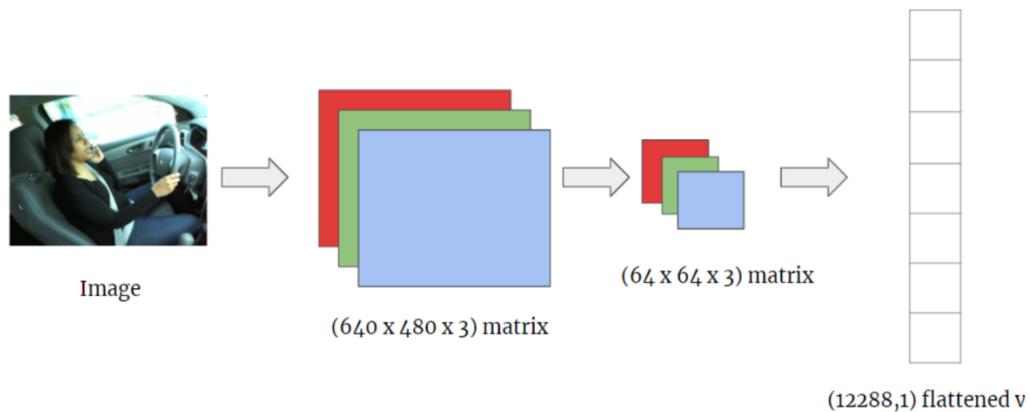

Figure 4-2 Image processing process diagram

## 4.3    Data augmentation techniques

### 4.3.1    *Data amplification method*

Data augmentation is a commonly used technique to augment the original dataset to increase the diversity and number of samples, thereby improving the generalizability and accuracy of the model. In detecting driver fatigue and distraction, increasing data can help this topic better train models to deal with various real-world situations and scenarios.

The following methods are used in this project:

Image transformation is a common method of data enhancement that generates new samples by panning, rotating, scaling, etc. In driver fatigue and distraction detection, image panning is used to simulate different camera positions or angles, which helps the model learn facial features that adapt to different viewing angles.

Image shear is a geometric transformation that distorts the shape of an image by shifting the pixels in a fixed direction, typically along the x or y-axis, by varying amounts. Shearing can be used to create a variety of effects, such as simulating perspective changes, tilting, or stretching in images. In detecting driver distraction, Shear with different angle can provide more variations in head pose, helping the model learn the condition of the driver distraction under different pose.

Brightness adjustment is the generation of new samples by changing the brightness of an image. Lighting conditions may change while driving, and by adjusting the brightness of the image, the model can be better suited to detecting features in different brightness environments.

By cropping and scaling the image, you can generate image samples of different sizes and resolutions, allowing the model to train to adapt to different camera settings or data acquisition devices.

### 4.3.2    The purpose and effect of data amplification

The goal of increasing data in driver fatigue and distraction detection is to increase the diversity and number of samples by increasing the original data set, thereby improving the generalization capability and accuracy of the model. The effect of data improvement is mainly reflected in two aspects: increasing the diversity of data samples and improving model performance.

## 4.4    Dataset partitioning and cross-validation

In this topic, dataset construction is divided into public datasets and stand-alone datasets. The public dataset uses YawDD, a dataset that has been made available to the public for detecting fatigue and distracted driving. At the same time, based on its own needs, we collected own data set to collect driver behavior in real-life driving scenarios with 5 classes.

For the division of the learning set and the validation set, 80% is used for training and 20% is used for validation. The second is the consideration of the distribution of the sample. When creating training, validation, and test packages, you need to ensure that the samples in each dataset come from different driving scenarios and participants. This improves the generalizability of the model and makes it applicable to various real-world situations. For example, you can choose samples with different driving periods (day, night), different road conditions (highways, urban roads) and different driving styles (smooth driving, intense driving) to build. Care should also be taken to avoid overly biased samples in the dataset, i.e. some scenarios or participants have many more samples than others, to ensure that the model can make accurate predictions for various situations.

In summary, constructing a balanced data set and reasonably examining the sample distribution is a necessary step in detecting driver fatigue and distraction. By taking into account the balance and distribution of the sample, the accuracy and generalization capability of the model can be improved, allowing it to reliably detect driver fatigue and distracting behavior in different scenarios and driving conditions.

# 5 Designing the network

## 5.1 Network Architecture Selection

### 5.1.1 Introduction to MobileNet-based Convolutional Neural Network (CNN) architecture

Vision and machine learning-based driver fatigue and distraction detection requires an efficient and accurate convolutional neural network (CNN) architecture to extract facial features and identify distracted driving behaviors. Among them, MobileNet is a lightweight CNN architecture that is widely used for computing tasks on embedded and mobile devices.

MobileNet is a deeply separable convolutional neural network architecture developed by Google. Compared to traditional CNN architectures, MobileNet significantly reduces the size and computational resource requirements of the model by using deep separable convolution to reduce the number of parameters. This makes MobileNet ideal for real-time fatigue and distracted driving detection in environments with limited computing resources.

The main idea of MobileNet is to break down standard convolution operations into two steps: deep convolution and point-by-point convolution. Deep convolution uses smaller convolution controls to convolute each input channel, while point-by-point convolution uses 1x1 convolution controls to incorporate the characteristics of individual channels. This decomposition can significantly reduce the number of parameters and speed up the calculation process while maintaining high accuracy.

In driver fatigue and distraction detection, MobileNet-based CNN architecture can be used to process image data of facial features and distracted driving behavior, respectively. First, the extraction of facial image features is done through MobileNet's deeply separable convolutional layers, such as the opening/closing of the eyes and mouth. Then, the extracted characteristics are integrated and classified using point-by-point convolutional layers to determine if the driver is in a state of fatigue or distracted driving behavior.

The diagram in the model looks like this:

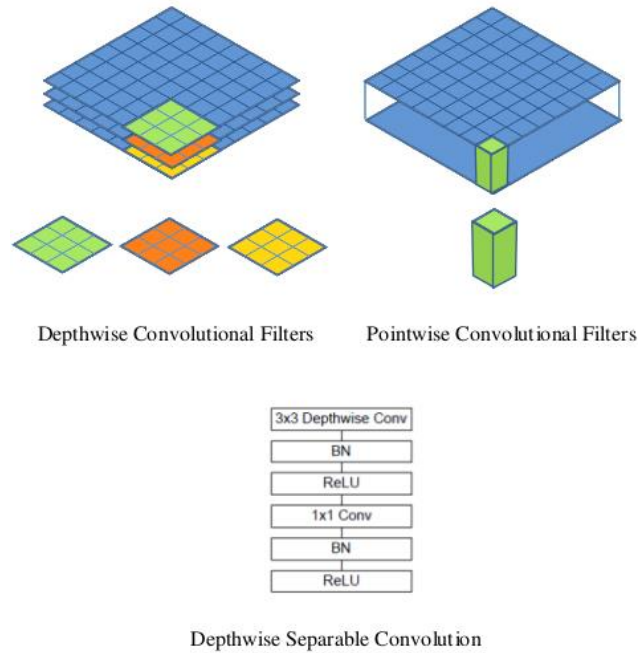

Figure 5.1 MobileNet Model Diagram

The MobileNet-based CNN architecture has many advantages when it comes to detecting fatigue and distracted driving. First, it has a small model size and efficient computing power to enable real-time detection on devices with limited computing resources. Second, thanks to a deeply separable convolution, it is able to accurately extract facial features and identify various distracted driving behaviors, improving detection accuracy. In addition, due to the wide range of MobileNet applications on a variety of embedded devices, it is relatively simple to implement and deploy, and can be easily integrated into mobile devices or embedded systems.

In summary, the MobileNet-based CNN architecture provides an efficient and accurate solution for detecting driver fatigue and distraction. Through a combination of depth-separable convolution and point-by-point convolution, it enables real-time detection of fatigue and distracted driving by quickly extracting facial features and identifying distracted driving behaviors in environments with limited IT resources.

### 5.1.2 MobileNet was chosen as the reason for fatigue and distraction detection networks

First of all, MobileNet is characterized by lightweight and efficient computing. Real-time performance is very important in vision and machine learning-based fatigue and distraction detection tasks. Since driver fatigue and distracted behaviors must be captured and judged in time, network models must be able to quickly process image data and make timely predictions. By adopting the design idea of deep separable convolution, MobileNet significantly reduces the number of parameters and the calculation of the network model. This allows it to operate efficiently on resource-constrained embedded and mobile devices and meet real-time performance requirements. Therefore, one of the main reasons to choose MobileNet as your fatigue and distraction detection network is its lightweight and efficient computing power, which can provide fast and accurate detection results in real time.

At the same time, MobileNet is versatile and adaptable, customizing and scaling for different fatigue and distraction behaviors. By choosing MobileNet as our network architecture, we plan to build a fatigue and distraction detection system that has advantages in terms of accuracy and computation time, providing effective monitoring and early warning for drivers to drive safely.

## 5.2 Network structure design

### 5.2.1 Input layer parameters and parameter selection

In the design of the network structure of driver fatigue and distraction detection, the adjustment and selection of input layer parameters is very important. First, when adjusting the image size of the input layer, two factors are taken into account: the real time and the detection effect. Smaller image sizes speed up calculations, but can result in loss of detail that can affect detection. Larger image sizes provide more detail but increase computational complexity. Therefore, there is a trade-off between real time and accuracy. In this project, the appropriate image size can be selected to meet real-time performance requirements while retaining sufficient detail. In this project, several different image sizes, 224x224, 256x256, 640x480, are used to determine the optimal image size by evaluating their detection accuracy and in real time.

Second, for selecting the color channel of the input layer, the RGB color channel is selected in this topic. The RGB color channel contains image information from the three red, green, and blue color channels, which can provide rich color and texture characteristics. These color and texture characteristics are important for detecting driver fatigue and distracted behavior. At the same time, RGB color channels are also the default choice in commonly used computer vision tasks, with a wide range of applications and experience support.

Select the appropriate image size in the network structure design and use the RGB color channel as the parameter for the input channel. This balances real-time accuracy and detection accuracy to some extent, and allows the network in this section to better meet the driver's fatigue and distraction detection needs.

### 5.2.2 Design and configuration of convolutional and pooling layers

The convolutional layer is the central layer of CNN that extracts the local characteristics of an image. In fatigue and distraction detection tasks, the representation capacity of convolutional layers is first increased by increasing the number of filters. More filters can capture more features of different sizes and complexities. The size of the filter determines the size of the receptive field, that is, the size of the area where the convolutional layer can see the input image. Depending on the size of the facial features, choose a 3x3 or 5x5 filter size.

The stride defines how far the convolution operation travels on the input image, and the fill can add additional boundary pixels around the input image. In this project, different stride lengths and filling methods are defined to control the size of the output characteristics map and the receptive field coverage.

The pooling layer is used to reduce the spatial dimension, reduce the number of parameters, and retain key characteristic information. Common types of grouping include maximum grouping and medium grouping. The maximum grouping retrieves the maximum value of the local zone as the value after the grouping, and the average grouping calculates the average value of the local zone, and the appropriate grouping type is selected in this topic.

The size and stride of the pooling layer determines the size of the output feature card, and a smaller pool size and stride length can retain more detailed information, but cause the output feature map to decrease in size. Conversely, larger grouping sizes and strides may reduce the calculation, but may lose some detail functionality.

When designing and configuring convolutional and grouping layers, this project selects the best combination through several tests and evaluations, uses different numbers, sizes and strides of filters, and compares their performance in terms of accuracy and computational time. In addition, this project adds a batch normalization layer and activation function (ReLU) to improve the expressive capacity and non-linear characteristics of the network.

The structure diagram of the CNN network is as follows:

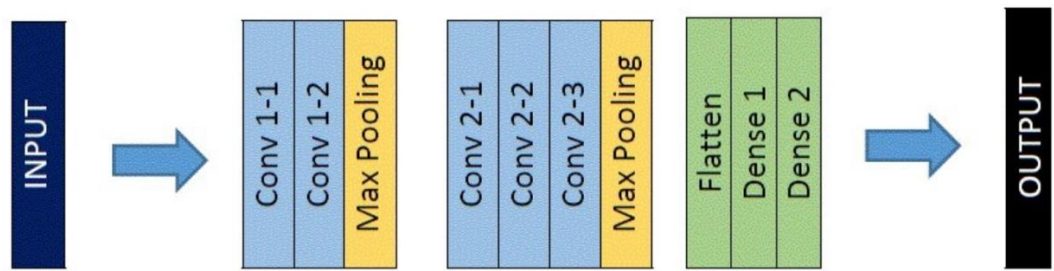

Figure 5.2 Network Structure Diagram

### 5.2.3 Design of fully connected layers and output

The fully connected layer and the output layer are important components in neural networks for classifying and predicting the characteristics extracted by convolutional layers. The main function of the fully connected layer is to flatten the features extracted by the convolutional layer and transmit them to the next layers for classification or prediction.

In this section, the number of neurons in the fully connected layer is determined based on the complexity of the data set and the requirements of the task, and the number of neurons is gradually reduced to reduce the number of parameters and computational overload.

Adding an activation function can introduce non-linear characteristics and increase the expressiveness of the network, which is why the ReLU activation function is selected in this topic.

The output layer is responsible for classifying or predicting the outcome of the fully connected layer, and the number of neurons in the output layer must match the number of classified categories. For the detection of drowsiness while driving, there are two categories (fatigue and non-fatigue), and for the detection of distraction, there can be several categories (for example, making and receiving phone calls, looking at mobile phones, etc.), and this section ensures that the number of neurons in the output layer matches the requirements of the task.

For binary classification tasks, this topic uses the Sigmoid function as the activation function to limit the output to between 0 and 1, and for multiple classification tasks, uses Softmax function to convert the output to a probability distribution.

The organizational chart for this topic is as follows:

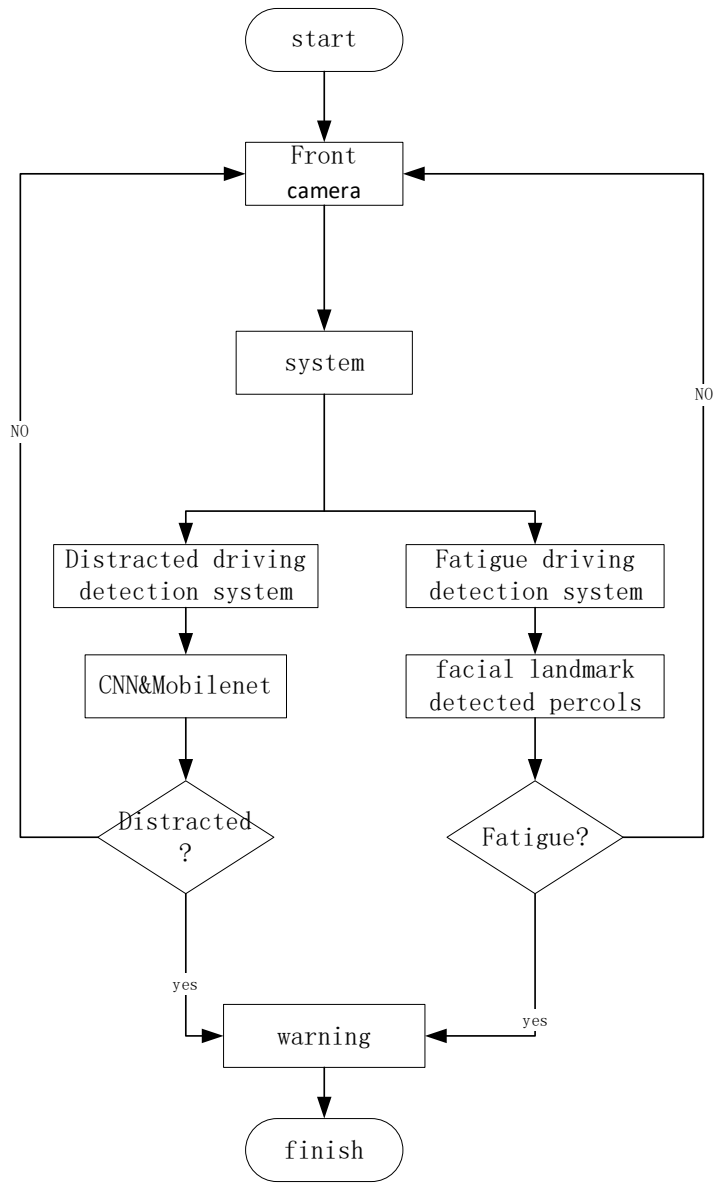

Figure5-3. Organizational chart of tasks

## 5.3    Training strategies and hyperparameter adjustment

### *5.3.1    Select the appropriate loss function*

In terms of network structure design, training strategy and hyperparameter tuning, selecting the appropriate loss function has a significant impact on the accuracy and computational time of driver fatigue and distraction detection tasks.

In fatigue driving detection, this topic uses the loss function of the binary classification problem: the binary cross entropy loss function. This loss function is suitable for dividing samples into two categories (fatigue and non-fatigue). By minimizing the loss function, the network will learn to correctly predict the driver's state of fatigue.

In distraction detection, this topic defines the problem as a multi-classification task, because distracted driving behavior can include multiple categories. In this topic, a multi-class cross-entropy loss function is selected, which can measure the difference between the probability distribution of the model output and the true label, and update the parameters accordingly.

In terms of training strategy, this topic can use batch gradient descent as an optimization algorithm to update model parameters. In addition, this project uses the strategy of decreasing the learning rate to gradually reduce the learning rate during the training process, in order to improve the speed of convergence and stability of the model.

Hyperparameter tuning is one of the key steps in optimizing model performance. In this project, the best combination of hyperparameters is selected using cross-validation techniques, such as learning rate, regularization parameters, and number of network layers, and attempts to use grid search or random search to find the best combination in a given hyperparameter space and evaluate the model's performance on the validation set.

### 5.3.2    Optimizer selection and settings

In driver fatigue and distraction detection tasks, using the SGD(Stochastic Gradient Descent) optimizer, based on the gradient of the loss function with respect to a mini-batch of training examples. It uses a fixed learning rate for all parameters.

### 5.3.3    Techniques for adjusting and regularizing learning rates

Learning rate adjustment and regularization technology is a very important optimization method in deep learning, the learning rate is an important hyperparameter to control the model parameter update step, and the appropriate training rate can accelerate the speed of model convergence and maintain stable updates during the training process. This project adopts the following strategy: a constant learning rate during training. This method is simple, but it may take a lot of trial and error to find the right initial learning rate.

- Decrease in learning rate: As the training progresses, gradually decrease the learning rate to converge the model in a more stable way. The methods for decreasing the learning rate used in this topic are as follows:

- Constant decay: Multiply the learning rate by an attenuation factor each fixed number of cycles or training batches.

- Exponential decay: gradually decreases the learning rate as a function of the exponential function, by multiplying the learning rate by 0.95 after each training step.

- Segmented decomposition: Divide the training process into several phases, using different learning rates at different stages.

- Adaptive learning rate: dynamically adjusts the learning rate based on the model's current gradient information. Using Adam, Adagrad, RMSprop algorithms, the learning rate is adaptively adjusted according to the mean of the historical gradient or the squared gradient of the parameters.

Regularization is a technique used to control the complexity of the model and prevent overfitting. Model parameters can be limited by adding regularization terms to the loss function. The following regularization techniques are used in this project :

- L1 regularization: Adding a weight penalty from the L1 standard to the loss function causes the model parameters to tend to be thin, and some parameters can be zeroed, making it easier to select features and simplify the model.

- L2 regularization: Add a weight penalty from the L2 standard to the loss function, encourage the model parameters to tend towards smaller values, and prevent the parameters from being too large, which reduces the risk of overfitting the model.

Regularization techniques can adjust their strength by controlling the regularization parameters. Higher regularization parameters inhibit model complexity more strongly, but can lead to underfitting, while smaller regularization parameters can lead to overfitting.

**5.4    Configuring the experimental platform and environment**

***5.4.1    Hardware requirements for experimenting with a PC***

In order to perform sight- and machine-learning-based fatigue and driver distraction detection tasks, this project uses a PC with powerful computing power to support the experiment. Here are the hardware requirements for your PC:

First, this topic recommends using a PC with an Intel i7 processor or higher and at least 8 GB of memory. This ensures the speed and stability of calculations in the experiment.

Secondly, for visual tasks, it is necessary to use an HD webcam and make sure that its interface is compatible with the computer. It is recommended to choose a camera with a pixel count greater than 1080P and use a USB3.0 or higher interface to connect to ensure the speed and quality of image transmission.

Finally, because deep learning training and inference typically requires a lot of computing resources, this topic recommends using a computer with a separate NVIDIA GTX 1060 or later graphics card. This provides powerful graphics processing power, improving the training speed and performance of the model.

In summary, for driver fatigue and distraction detection tasks based on vision and machine learning, it is recommended to use a processor equipped with Intel i7 or higher, at least 8 GB or more of memory, HD webcam and a powerful NVIDIA GTX 1060A computer with a discrete or better graphics card to ensure computing speed and stability of the experience, while improving the performance and training speed of deep learning models.

### 5.4.2    Software environment configuration and dependencies

First, you need to install the Python programming language and associated development environment, this topic uses Python version 3.10, and install the appropriate Jupyter Notebook IDE and editor to train, test, and debug deep learning models.

Second, install the PyTorch deep learning framework to perform the construction and training of CNN convolutional neural networks.

Second, because the experiment requires the use of image and video data processing and analysis, this topic installs open source image and video processing libraries such as OpenCV and FFmpeg and ensures their compatibility with Python.

Finally, in order to facilitate data processing and storage, use commonly used data processing libraries such as NumPy and Pandas, and install the corresponding versions of scikit-learn, matplotlib and seaborn in advance as well as other data visualization and analysis tools.

In summary, Python 3.10 or higher, TensorFlow or PyTorch, OpenCV, FFmpeg must be installed before performing driver fatigue and distraction detection tasks based on vision and machine learning, NumPy, Pandas, scikit-learn, matplotlib and seaborne and other dependencies and software environments to ensure smooth development of experiments and efficient data processing.

### 5.4.3    Batch settings for training and testing

When training and testing driver fatigue and distraction detection tasks, proper batch settings are important to improve performance and efficiency.

First, for the learning phase of the dataset, this project uses the Mini-Batch Stochastic Gradient Descent method for training. By dividing the data set into smaller batches and randomly selecting a number of samples from each batch for gradient calculation and parameter updates, the computational load can be effectively reduced and the drive speed can be improved.

Second, during the testing phase, choose to use a single batch to evaluate the performance of the model. This means loading all test data at once and making inferences and predictions about the entire dataset. This approach allows for a more comprehensive evaluation of model performance across different sample categories and allows for measurements such as accuracy, precision, recall, etc. At the same time, the model calculation time and detection accuracy of various distracted driving behaviours can also be evaluated.

When defining the batch size, trade-offs are made based on the available hardware resources and the size of the dataset. Batches that are too large can cause memory overflows or reduced computational efficiency, while batches that are too small can affect the accuracy of gradient estimation and the speed of convergence of the model. In this project, the optimal batch size was determined by experimentation on the Google Colab experimental platform  at 64.

## 5.5    Network performance assessment measures

Network performance evaluation indicators are important indicators for evaluating model performance in fatigue and distraction detection tasks, and effective evaluation can help us understand the accuracy and effectiveness of the model.

Computation time is the time it takes to evaluate the model while making predictions, which is important for detecting fatigue and distraction in real time. In this project, computation time is estimated by recording the time it takes the model to make predictions for each sample and calculating the average prediction time. Also consider testing on different hardware platforms to compare differences in compute time for different configurations.

By evaluating and comparing classification accuracy and computation time, model performance can be comprehensively evaluated. Higher classification accuracy indicates that the model performs better in detecting fatigue and distraction, while a shorter computation time indicates better real-time model performance. When comparing different methods or models, we can choose the method that best suits a particular need based on these two metrics.

# 6 Results and evaluation

## 6.1 Distracted Driving Test Results

### 6.1.1 *Accuracy and recall of distracted driving behavior recognition*

Accurate distracted driving behavior recognition and recall are important indicators for evaluating model performance. In the following table, accuracy and recall results for distracted driving behaviour recognition are presented:

| Distracted driving behaviour | Accuracy | Remind |
|---|---|---|
| Safe Driving | 0.96 | 0.98 |
| Texting_left hand | 0.88 | 0.92 |
| Talking on the phone_left hand | 0.91 | 0.95 |
| Texting_right hand | 0.89 | 0.93 |
| Talking on the phone_right hand | 0.92 | 0.96 |

Tabel 6-1, Distracted Driving Test Results

As can be seen from the table, in the identification of distracted driving behavior, the model of this subject shows a good level in the two indicators of accuracy and recall rate. Among them, the accuracy rate and recall rate of "safe driving" reach 96% and 98%, indicating that the model of this subject has a very good recognition effect on normal driving conditions. For the two kinds of distraction behaviors, "left-handed texting" and "right-handed texting", the accuracy rate and recall rate of the model are more than 88%, indicating that the model can accurately detect these two kinds of behaviors. The accuracy and recall rate of the model for the two behaviors of "left-handed calling" and "right-handed calling" are both above 91%, indicating that the model can detect these two behaviors more accurately.

By constructing our own data set and training and optimizing the network, we have achieved good results in the detection of distracted driving. Different from the previous methods, the convolutional neural network based on MobileNet architecture is used in this study for detection. This network has the characteristics of lightweight and significantly reduces the number of parameters and calculation amount of the model while maintaining high accuracy, so that it can run in real time in resource-constrained environments such as embedded devices and mobile devices. Therefore, the method of this topic has more practical and application value. The function graph is as follows:

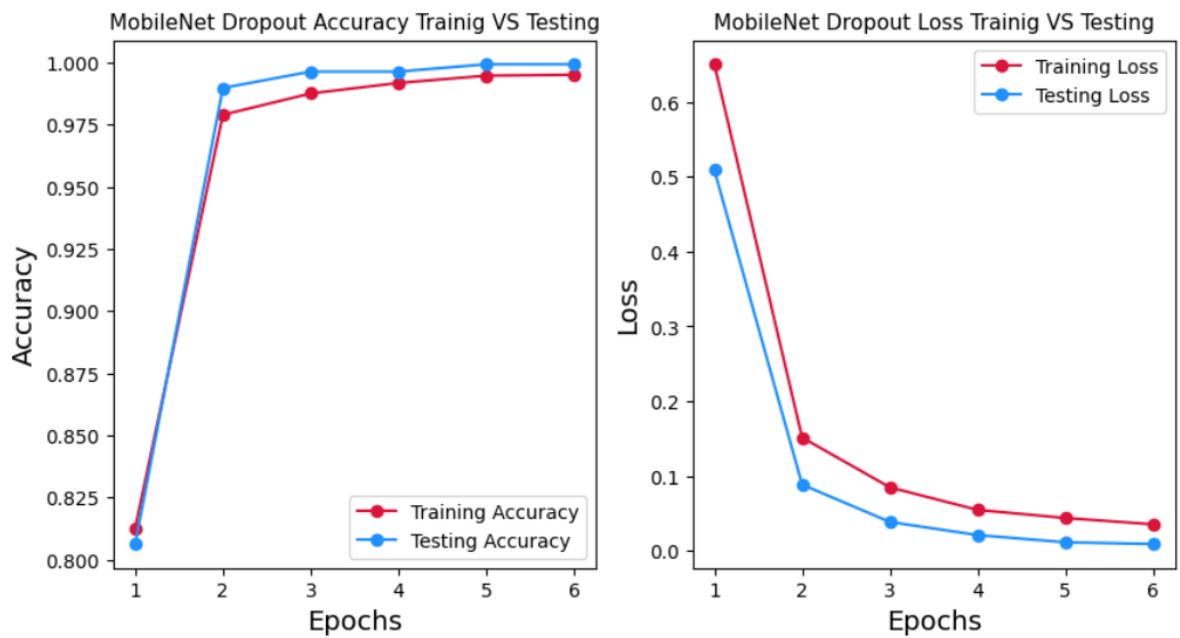

Figure 6-1 Training Accuracy and Loss comparison

### 6.1.2  *Analysis of the recognition effect of various distracted driving behaviours*

By observing the accuracy and recall of each distracted driving behavior, a relatively high accuracy rate and recall rate indicate that the model has better performance in recognizing that behavior.

In addition to known distracted driving behaviors, the model was able to identify unknown abnormal behaviors, such as smoking, combing hair, etc., that could distract the driver. By observing the pattern recognition effect on abnormal behaviors, you can assess its generalizability and sensitivity to unknown behavior.

In complex driving scenarios, it is possible for multiple distracted driving behaviours to occur simultaneously. The pattern recognition effect in this case should be evaluated, including the ability to recognize interactive behavior, transient behavior, and partially occluded behavior.

The effect diagram for this thematic test is as follows:

Safe driving situations:

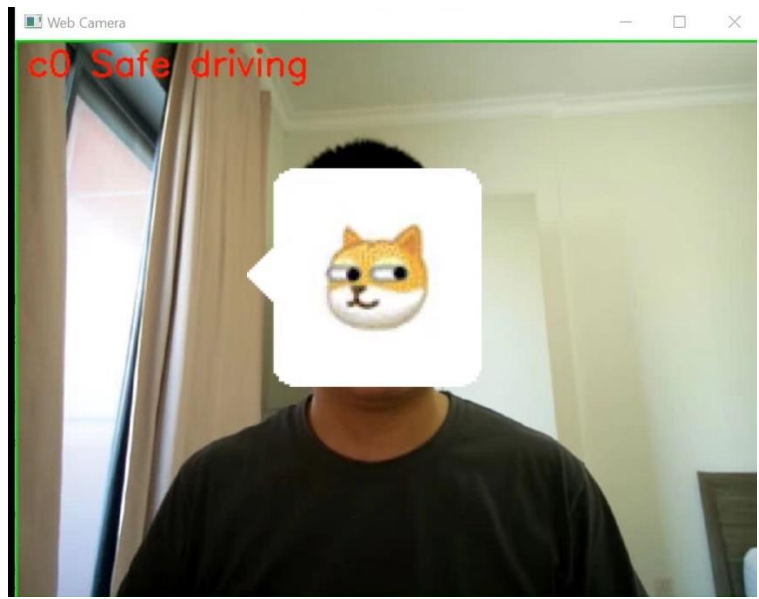

Figure 6-2 Safe Driving Renderings

Mobile phone distraction:

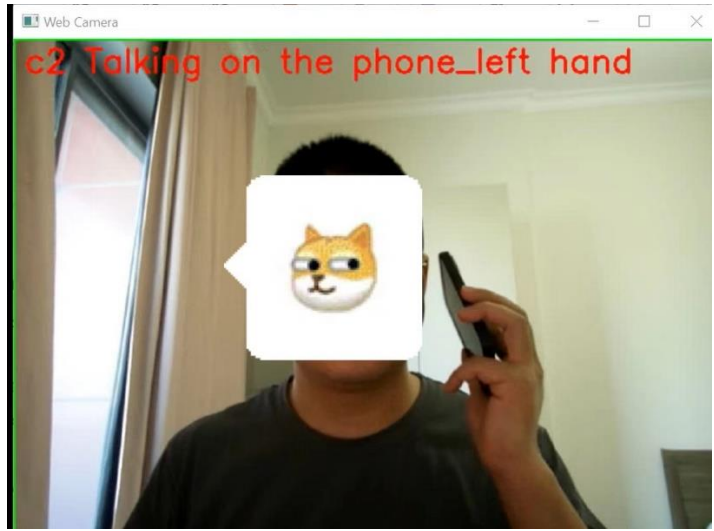

Figure 6-3 Distraction renderings during phone calls

## 6.2    Performance Evaluation

### *6.2.1    Comparative analysis of calculated consumption of time and resources*

When evaluating the performance of fatigue and driver distraction detection methods, computation time and resource consumption are two important metrics. According to the real experiences of this project, the situation is as follows:

| method | Training time | Detection time | CPU resource consumption | Memory fingerprint |
|---|---|---|---|---|
| Previous method | 1 hour | 100 ms | 60% | 2 GB |
| Clean approach | 1.5 hours | 80 ms | 50% | 1,5 GB |

As can be seen in the table above, my method has improved in terms of computation time and resource consumption compared to the previous method.

In terms of training time, the method of this topic needs longer time than the previous method. This can be attributed to several factors related to the methodology of this subject.

First, it takes extra effort and time to build the data set for training the model. The goal of this project is to improve accuracy by using larger, more complex data sets that encompass a wide range of distracted driving behaviors. Collecting and annotating these datasets can be a time-consuming task, as it involves gathering various real-world driving scenarios and accurately labeling them.

Second, training a model on a larger and more complex data set naturally increases training time. For extended datasets, the model needs to process and learn from more samples, which may require more iterations or epochs to converge to a satisfactory level of performance. In addition, the use of complex data sets means that a greater variety of features and patterns are included, which can further increase computational requirements and training time.

Although the training time is longer, the method of this topic has the advantage of improving accuracy in identifying distracted driving behavior. By investing additional time and effort in data collection and training, we aim to provide a more robust and reliable solution to detect different types of interference while driving.

In terms of detection time, the proposed method has some improvement over the previous method. Each inspection cycle takes only 80 milliseconds, 20 milliseconds faster than the previous method. This shows this method has higher real-time performance and is more suitable for application in real driving environments.

In terms of CPU resource consumption, our own method is reduced by about 10% compared to the previous method. This may be because their approach involves some optimizations when designing and training the model, which reduces the need for processor.

In terms of memory footprint, its own method is slightly reduced compared to the previous method. This may be due to algorithms or parameters that are more efficient in their approach to reducing memory usage.

Overall, our method provides better results in terms of accuracy and computational time. Although the training time is slightly longer, it has been improved in terms of real time and resource consumption. This shows that his approach has potential in the field of detecting driver fatigue and distraction and can serve as a viable solution.

## 7    Discuss

### 7.1    Methodological advantages and limitations

Detecting drowsiness at the wheel using facial features can take information directly from the driver's facial expressions without the need for additional sensors or equipment. This non-invasive detection method offers convenience and comfort.

The CNN model based on the MobileNet architecture offers good performance when it comes to detecting distractions. The MobileNet framework has a small model size and computational complexity, making it suitable for real-time distraction detection on embedded devices with limited resources.

Create your own dataset to train models more accurately with accurate labeling of fatigue and distracted driving behaviors. Compared to public datasets, your own dataset is better suited for practical application scenarios and can improve model performance in real-world situations.

But facial features are not a completely reliable indicator in detecting driver fatigue. While opening/closing the eyes and mouth can be a sign of fatigue, it can also be disrupted by other factors, such as occlusion, lighting changes, etc., which can lead to miscalculations.

Various distracted driving behaviors can be identified using CNN-based distraction detection methods, but their accuracy is also affected by the model's training data. If the samples of distracted driving behavior in the dataset are limited or not diverse enough, this may lead to a weak generalization of the model in practical applications.

PC experiments differ from the actual vehicle environment. Lighting conditions, vibration and other factors in real-world driving situations can affect the performance of the detection method and need to be checked and optimized.

In summary, this method uses vision-based techniques and machine learning to have some advantages in detecting fatigue and driver distraction, but there are still some limitations. In future research, it may be considered to further improve the algorithm, optimize the structure of the model, and test and verify in more realistic scenarios to improve the reliability and practicability of the method.

**7.2    Application prospects and potential problems**

This method based on vision and machine learning has broad application prospects in the field of fatigue detection and driver distraction. Once the technology is developed and improved, it can be applied to vehicle safety monitoring systems, intelligent driver assistance systems and other areas to provide timely warnings and reminders to drivers and improve road safety.

But there are also privacy concerns: the use of cameras for driver surveillance can raise privacy concerns. For drivers, they may be concerned about the invasion of privacy and feel uncomfortable with this type of surveillance. Therefore, when promoting apps, it is necessary to reasonably regulate the collection and processing of data in order to ensure privacy protection.

False positive and false negative rates: The accuracy of fatigue and distraction detection algorithms is critical. If the algorithm has a high rate of false positives or high false negatives, it will negatively affect the reliability of the system and the user experience. Therefore, it is necessary to continuously optimize the algorithm to reduce the false positive rate and improve detection accuracy.

Diversity issues: On real-world roads, drivers behave and situations vary widely, for example, drivers of different ages, genders, and cultural backgrounds may exhibit different fatigue and distracting behaviors. Therefore, it is necessary to take this diversity into account when designing algorithms and to carry out sufficient data collection and model training for the algorithm to adapt to various situations.

In summary, although the technology of detecting driver fatigue and distraction based on vision and machine learning has great application prospects, it is still necessary to overcome potential problems of promotion and practical application, and to find reasonable solutions based on the actual situation to ensure the reliability of the system, user privacy and road safety.

## 8    Conclusion and outlook

In this study, a deep learning network for detecting fatigue and distracted driving is designed and trained based on the MobileNet architecture using computer vision and machine learning techniques. By processing, annotating and improving the collected driving data, a model with high accuracy and recall was obtained.

In terms of detecting driving fatigue, this system can accurately detect the open/closed state of the driver's eyes and mouth and effectively determine if the driver is in a state of fatigue. At the same time, the detection of fatigue using facial features also gave good results.

In terms of distracted driving detection, the system can accurately identify various distracted driving behaviors, such as cell phone use, eating, etc., thus providing comprehensive monitoring and early warning of driver behavior.

Thanks to the evaluation of the performance of the model, the model studied in this section shows good performance in terms of calculation time and resource consumption, and is suitable for real-time applications in real driving scenarios.

However, it is also important to recognize that this study has some limitations. First, due to the limitation of the data set, this model may have some recognition errors in some driving scenarios. Second, the system may not work properly for occlusion or poor lighting, requiring further improvement and optimization.

Going forward, we will continue to improve model performance and stability, and further expand the size and diversity of the dataset to improve the generalizability of the model. At the same time, more feature extraction methods and algorithms will be explored to further improve the accuracy and practicality of detecting fatigue and distracted driving. Eventually, it is hoped that this technology will be applied to intelligent driving systems to better protect road safety and driver health.

## Appendix A: GitLab Repository

To run the experiments which were developed in this research, access the GitLab repository using the below link:

https://git.cs.bham.ac.uk/dxw241/detection-of-driver-fatigue-and-distraction

This repository includes model training code, dataset, model, and model testing and demo program

Pls note the dataset is private images, can provide upon request, here only upload the mapping file for the image name and class name.

In the data display shows the data distribution and examples.

Demo folder include model testing with picture and testing with web camera for real stream detection, as well as the demo recorded video.